\documentclass{article}
\usepackage{amsmath,amssymb}
\usepackage{bm} 
\usepackage{arxiv}
\usepackage[utf8]{inputenc} 
\usepackage[T1]{fontenc}    
\usepackage{hyperref}       
\usepackage{url}            
\usepackage{booktabs}       
\usepackage{amsfonts}       
\usepackage{nicefrac}       
\usepackage{microtype}      
\usepackage{lipsum}
\usepackage{graphicx}
\graphicspath{ {./images/} }
\usepackage[numbers]{natbib}
\usepackage{subcaption}
\usepackage{float}
\usepackage{tabto}
\newcommand{\E}{\bm{E}}
\newcommand{\Hh}{\bm{H}}
\newcommand{\W}{\bm{W}}

\title{Fusion Matters: Length-Aware Analysis of Positional-Encoding Fusion in Transformers}

\author{
 Mohamed Amine Hallam \\
  School of Computer Science and Technology\\
  Harbin Institute of Technology\\
  \texttt{25sf51027@stu.hit.edu.cn} \\
   \And
 Kuo-Kun Tseng \\
   School of Computer Science and Technology\\
   Harbin Institute of Technology\\
   \texttt{kktseng@hit.edu.cn} \\
}

\begin{document}
\maketitle
\begin{abstract}
Transformers require positional encodings to represent sequence order, yet most prior work focuses on designing new positional encodings rather than examining how positional information is fused with token embeddings. In this paper, we study whether the fusion mechanism itself affects performance, particularly in long-sequence settings. We conduct a controlled empirical study comparing three canonical fusion strategies—element-wise addition, concatenation with projection, and scalar gated fusion—under identical Transformer architectures, data splits, and random seeds. Experiments on three text classification datasets spanning short (AG News), medium (IMDB), and long (ArXiv) sequences show that fusion choice has negligible impact on short texts but produces consistent gains on long documents. To verify that these gains are structural rather than stochastic, we perform paired-seed analysis and cross-dataset comparison across short, medium, and long sequence regimes. Additional experiments on the ArXiv dataset indicate that the benefit of learnable fusion generalizes across multiple positional encoding families. Finally, we explore a lightweight convolutional gating mechanism that introduces local inductive bias at the fusion level, evaluated on long documents only. Our results indicate that positional-encoding fusion is a non-trivial design choice for long-sequence Transformers and should be treated as an explicit modeling decision rather than a fixed default.
\end{abstract}

\section{Introduction}
Transformers have become the dominant architecture for sequence modeling across natural language processing tasks, including text classification, machine translation, and long-document understanding. Because self-attention is inherently permutation-invariant, positional information must be injected explicitly in order to represent token order. As a result, positional encodings have been a core component of Transformer architectures since their introduction.\cite{vaswani2017attention}

A substantial body of prior work has focused on \textit{what} positional information should be encoded. This includes sinusoidal encodings, learned absolute embeddings\cite{vaswani2017attention}, relative position representations\cite{shaw2018relative}, rotary embeddings\cite{su2021roformer}, and bias-based formulations\cite{press2022alibi}. These approaches differ in how they represent order, distance, or relative position, and many have been proposed specifically to address limitations of Transformers on long sequences.

However, an orthogonal and largely overlooked design choice concerns \textit{how} positional encodings are fused with token embeddings. In most standard implementations, positional encodings are injected via simple element-wise addition. This practice implicitly assumes that positional information should contribute uniformly across tokens and layers, and that its influence should be fixed rather than learned. While this assumption has become a de-facto default, it is rarely justified explicitly and is seldom questioned in empirical studies.

This omission is notable given recent trends in long-context modeling. Modern benchmarks increasingly involve long documents, such as scientific articles, legal texts, and multi-paragraph narratives, where positional relevance may vary substantially across the sequence. In such settings, treating positional information as a uniform additive signal may be unnecessarily restrictive. Yet, despite extensive work on alternative positional encoding \textit{representations}, the fusion mechanism itself is typically held fixed.\cite{tay2021lra,beltagy2020longformer}

In this work, we isolate the positional-encoding fusion operator as an independent modeling variable. Rather than proposing a new positional encoding, we ask whether the \textit{method used to combine} token embeddings and positional encodings materially affects Transformer performance, particularly as sequence length increases. Our focus is intentionally narrow: we hold the Transformer architecture, optimization procedure, data splits, and random seeds constant, and vary only the fusion mechanism.

We investigate three core questions:

\begin{enumerate}
    \item \textbf{Does the choice of positional-encoding fusion operator affect downstream performance when all other factors are controlled?}
    \item \textbf{Is any observed effect dependent on sequence length, or does it persist across short, medium, and long documents?}
    \item \textbf{Can inductive bias be introduced at the fusion level—without modifying attention or the Transformer encoder—to better handle long sequences?}
\end{enumerate}
To answer these questions, we conduct a controlled empirical study across three text classification datasets spanning short (AG News), medium (IMDB), and long (ArXiv) sequence regimes. We compare element-wise addition, concatenation with projection, and learnable gating under identical experimental conditions. To rule out stochastic explanations, we perform paired-seed analyses in which fusion methods are compared using identical random seeds and data splits.

Beyond global gating, we further explore a lightweight convolutional gating mechanism applied solely to positional encodings. This design introduces local inductive bias at the fusion stage while leaving the Transformer core unchanged. The goal is not to outperform specialized long-sequence architectures, but to assess whether modest structural bias at the fusion level can yield consistent gains on long documents.

Our results show that positional-encoding fusion is not a neutral design choice. While fusion differences are negligible on short and medium-length datasets—where performance saturates—they produce consistent and statistically stable improvements on long documents. These findings suggest that positional-encoding fusion should be treated as an explicit modeling decision rather than a fixed default. Across long-document classification on ArXiv, scalar gated fusion improves accuracy by approximately 6.5 absolute points over standard additive fusion, while producing no consistent gains on short or medium-length datasets.

\textbf{Contributions.}

 This paper makes the following contributions:

\begin{itemize}
    \item We provide the first controlled, cross-dataset study isolating positional-encoding fusion as a modeling variable in Transformers.
    \item We show that learnable fusion mechanisms yield consistent improvements on long-sequence classification, while offering little benefit on short or medium-length texts.
    \item We introduce and evaluate a lightweight convolutional gating mechanism that injects local inductive bias at the fusion level without modifying attention.
    \item We demonstrate robustness of these findings through paired-seed analysis and evaluation across multiple positional encoding families.
\end{itemize}

\section{Related Work}

The original Transformer introduces sinusoidal positional encodings that are combined with token embeddings through element-wise addition \cite{vaswani2017attention}. Subsequent work explored alternative representations, including learned absolute embeddings, relative position representations \cite{shaw2018relative,dai2019transformerxl,he2021deberta}, rotary embeddings \cite{su2021roformer}, and bias-based approaches such as ALiBi \cite{press2022alibi}. Variants of relative positional bias have also been adopted in large-scale pretrained models, such as the T5 architecture \cite{raffel2020exploring}. Several studies analyze the limitations and extrapolation behavior of existing positional encodings \cite{ke2021rethinking,chi2022position}, highlighting that the choice of positional representation can significantly affect long-context modeling.\\
A separate line of research addresses the challenge of modeling long sequences by modifying attention patterns or improving computational efficiency. Sparse-attention architectures such as Longformer \cite{beltagy2020longformer} and BigBird \cite{zaheer2020bigbird} reduce the quadratic complexity of self-attention, while approximate or low-rank methods such as Reformer \cite{kitaev2020reformer}, Performer \cite{choromanski2021performer}, and Linformer \cite{wang2020linformer} enable linear-time attention. Hierarchical encoders such as ETC further target structured long inputs \cite{ainslie2020etc}. While effective, these approaches typically introduce architectural changes and retain the standard additive fusion of token embeddings and positional encodings.\\
Gating and modulation mechanisms have been explored within Transformer architectures to control information flow in attention and feed-forward layers. Examples include gated linear units \cite{shazeer2020glu}, stabilization techniques for training Transformers \cite{parisotto2020stabilizing}, and feature-wise modulation mechanisms such as FiLM \cite{perez2018film}. Related ideas have also been applied to positional signals within attention \cite{he2021deberta}. However, these mechanisms are generally applied to intermediate representations rather than to the fusion of token embeddings and positional encodings at the model input.\\
Across most Transformer variants, the fusion of token embeddings and positional encodings is inherited from the original architecture and implemented as a fixed additive operation \cite{vaswani2017attention}. Even when alternative positional representations or attention mechanisms are introduced, the fusion operator itself is rarely varied or analyzed in isolation. In contrast, our work focuses on this under-explored design dimension by systematically evaluating different fusion operators while holding architecture, optimization, and data splits fixed across sequence-length regimes.

\section{Method}

\subsection{Base Model and Positional Encoding}
All experiments use a fixed encoder-only Transformer architecture. To avoid confounding factors, the architecture, optimization procedure, and training schedule are identical across fusion variants.

We use\textbf{ sinusoidal positional encodings} throughout the main experiments. This choice enables direct comparison across datasets of different lengths and avoids introducing additional variability from alternative encoding schemes.
\subsection{Positional-Encoding Fusion Operators}
Let $\E \in \mathbb{R}^{L\times d}$ denote token embeddings and $\bm{P} \in \mathbb{R}^{L\times d}$ positional encodings, where $L$ is sequence length and $d$ is model dimension.
\subsubsection{Additive fusion (Add)}
The standard approach combines token embeddings and positional encodings via element-wise addition:
\begin{equation}
\Hh = \E + \bm{P}.
\end{equation}
This method does not introduce additional parameters and assumes a fixed, uniform contribution of positional information across tokens.
\subsubsection{Concatenation with projection (Concat).}
Token embeddings and positional encodings are concatenated and projected back to the model dimension: 
\begin{equation}
\Hh = \W \,[\E;\bm{P}], \qquad \W \in \mathbb{R}^{d \times 2d},
\end{equation}
Where [E;P] denotes concatenation along the feature dimension. This operator allows the model to learn a linear combination of content and positional features.
\subsubsection{Gated Addition (Gate-Scalar).}
To relax the assumption that positional information should contribute uniformly across all tokens, we introduce a \emph{gated fusion} mechanism that learns a position-dependent trade-off between token embeddings and positional encodings.

Let $\mathbf{E}\in\mathbb{R}^{L\times d}$ denote token embeddings and $\mathbf{P}\in\mathbb{R}^{L\times d}$ positional encodings. For each token position $i$, we compute a scalar gate based on the concatenation of token and positional representations:
\begin{equation}
g_i = \sigma\!\left(\mathbf{w}^\top [\mathbf{E}_i;\mathbf{P}_i] + b\right),
\end{equation}
We refer to this variant as \textit{Gate-Scalar} to distinguish it from more expressive MLP-based gating mechanisms evaluated separately in Supplementary Experiments \ref{sec:Supplementary Experiments}. 
where $[\mathbf{E}_i;\mathbf{P}_i]$ denotes feature-wise concatenation, $\mathbf{w}\in\mathbb{R}^{2d}$ and $b\in\mathbb{R}$ are learnable parameters, and $\sigma(\cdot)$ is the sigmoid function. \\In all main experiments, the gate is a scalar computed from the concatenation of token and positional representations, This design introduces minimal interaction, ensuring that fusion remains a lightweight modulation rather than a full content–position transformation. 

The gate is a scalar value computed independently at each position and shared across feature dimensions.
\begin{equation}
\mathbf{H}_i = g_i\,\mathbf{E}_i + (1-g_i)\,\mathbf{P}_i.
\end{equation}
\begin{figure}[H]
  \centering
  \begin{subfigure}{0.45\textwidth}
    \centering
    \includegraphics[width=\linewidth]{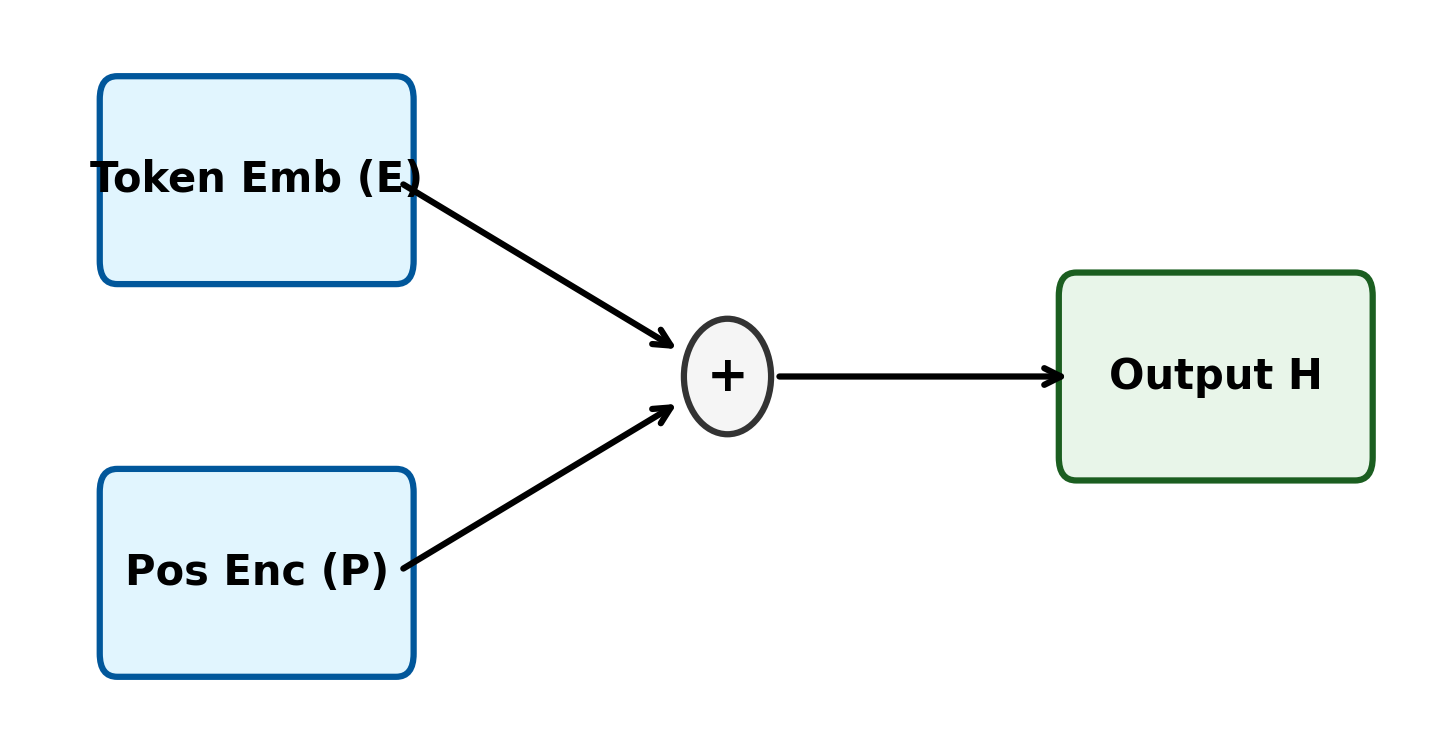}
    \caption{Additive (Add)}
    \label{fig:add}
  \end{subfigure}\hfill
  \begin{subfigure}{0.45\textwidth}
    \centering
    \includegraphics[width=\linewidth]{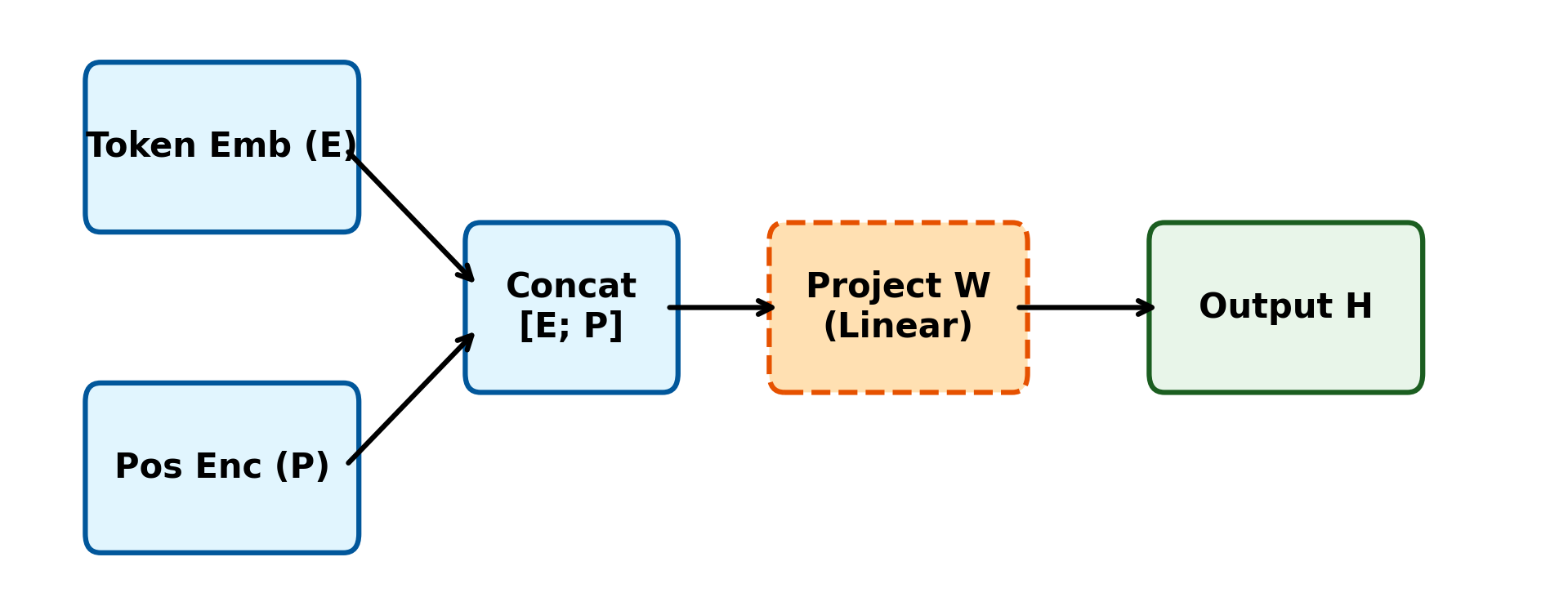}
    \caption{Concat + projection (Concat)}
    \label{fig:concat}
  \end{subfigure}\hfill

  \caption{Fusion operators in the evaluated experiments.}
  \label{fig:fusions-main}
\end{figure}

\subsection{Local Convolutional Positional Gating}

In addition to global gated fusion, we explore a local convolutional gating variant evaluated on long documents only. The motivation is to introduce a mild inductive bias that allows the contribution of positional information at a given token to depend on its local neighborhood, without modifying the Transformer encoder or attention mechanism. 
We instantiate this local gating using a depth-wise 1D convolution; other local sequence models (e.g., TCNs or shallow MLPs over local windows) are possible but not explored here.

For each token position $i$, we compute a scalar gate by applying a depth-wise one-dimensional convolution over the positional encodings:
\begin{equation}
g_i = \sigma\!\left( \sum_{k=-K}^{K} \mathbf{w}_k \odot \mathbf{P}_{i+k} \right),
\end{equation}
where $\{\mathbf{w}_k\}$ are learnable depth-wise convolution kernels of size $2K+1$, $\sigma(\cdot)$ denotes the sigmoid function, and $\odot$ represents element-wise multiplication.

The fused representation is then obtained using the same convex combination as in global gated fusion:
\begin{equation}
\mathbf{H}_i = g_i\,\mathbf{E}_i + (1-g_i)\,\mathbf{P}_i.
\end{equation}

Unlike global scalar gating (Gate-Scalar), which determines the gate independently at each position, this formulation allows positional relevance to be modulated based on nearby positions. Importantly, the convolution operates solely on the positional encodings and introduces no changes to the self-attention mechanism or Transformer architecture. We evaluate this variant as an exploratory design to assess whether locality-aware fusion provides additional benefits in long-sequence settings.
\begin{figure}[H]
  \centering
  
  \includegraphics[width=0.75\linewidth]{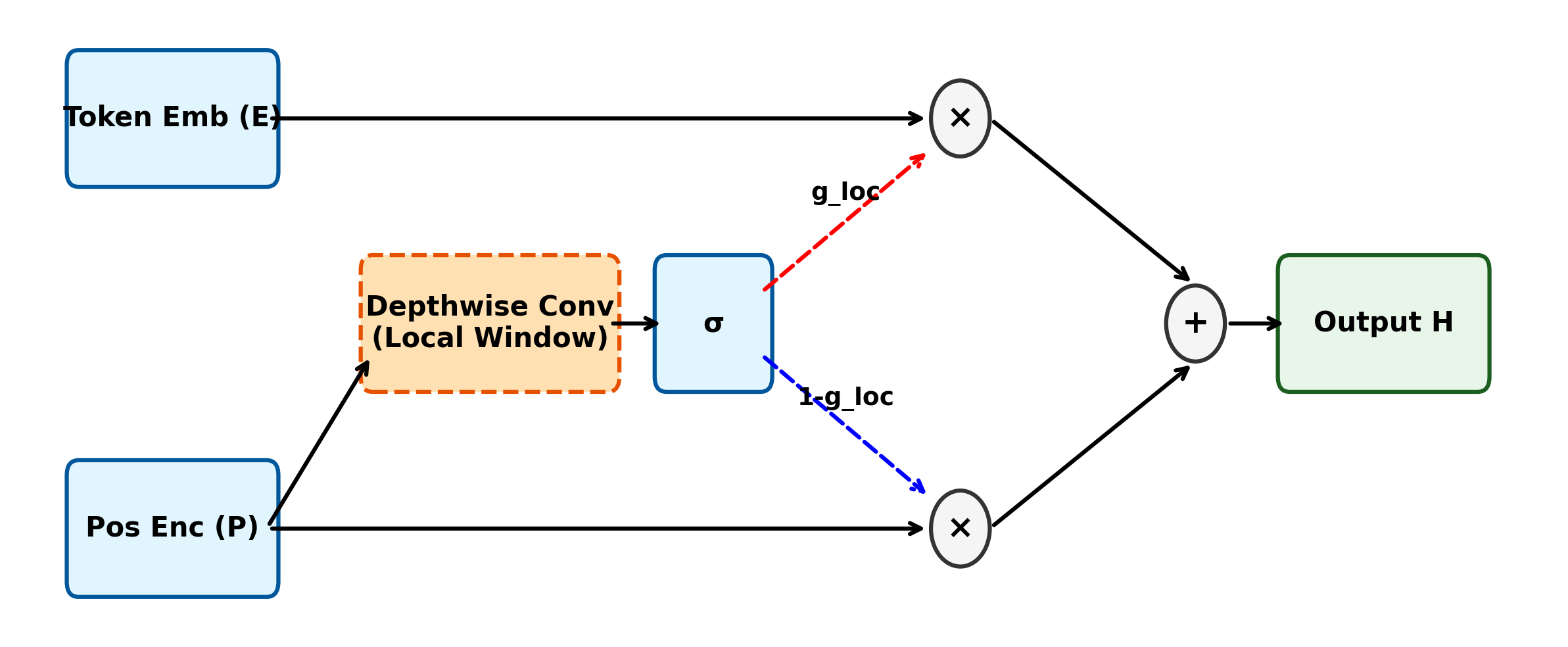}
  \caption{Local convolutional positional gating (Gate-CNN). The gate is computed from a depth-wise 1D convolution over positional encodings only, then used to mix token and positional features.}
  \label{fig:gate-cnn}
\end{figure}

\section{Experimental Setup}
\subsection{Dataset}
We evaluate on three text classification datasets with increasing sequence length:\\
\tabto{1cm} 1. Ag News (Short sequences)\\
\tabto{1cm} 2. IMDB (medium-length sequences)\\
\tabto{1cm} 3. ArXiv (long documents, up to several thousand tokens)\\
We focus on classification tasks because they isolate sequence representation quality and avoid confounding factors introduced by decoding or generation.
\subsection{Protocol}
Across all experiments, data splits and hyperparameters are held fixed across fusion variants, and models are trained using multiple random seeds. For paired comparisons, the same random seed is used across fusion methods, ensuring identical initialization and data ordering. This experimental protocol isolates the effect of the fusion operator and ensures that observed performance differences cannot be attributed to stochastic variation. All fusion operators differ only in the embedding fusion step; the Transformer encoder and attention mechanism are unchanged. Code and reproduction materials are available online \cite{hallam2026fusionmatters}.
\section{Fusion Benchmark Lengths}
\begin{table}[h]
    \centering
    \caption{Mean test accuracy (± standard deviation) across fusion strategies using sinusoidal positional encodings}
    \label{tab:fusion_results}
    \begin{tabular}{lccc}
        \toprule
        Dataset & Add & Concat & Gate-Scalar\\
        \midrule
        \textbf{AG News} & 91.15 $\pm$ 0.08 & 90.93 $\pm$ 0.11 & 91.07 $\pm$ 0.09 \\
        \textbf{IMDB}   & 83.28 $\pm$ 0.15 & 83.78 $\pm$ 0.13 & 83.40 $\pm$ 0.14 \\
        \textbf{ArXiv}  & 59.22 $\pm$ 0.32 & 63.44 $\pm$ 0.28 & \textbf{65.73 $\pm$ 0.30} \\
        \bottomrule
    \end{tabular}
\end{table}
On AG News and IMDB, performance differences between fusion methods are small and inconsistent, indicating rapid saturation on short and medium-length sequences. In contrast, on ArXiv, Gate-Scalar consistently outperforms additive fusion, yielding a substantial absolute improvement in accuracy. A more expressive MLP-based fusion baseline is evaluated separately on the ArXiv dataset and reported in the \textbf{(Supplementary Experiment~\ref{sec:Supplementary Experiments})}. 
\section{Robustness Across Sequence-Length Regimes}
\subsection{Paired-Seed Delta Analysis}
\begin{figure}[H]
  \centering
  \includegraphics[width=0.55\linewidth]{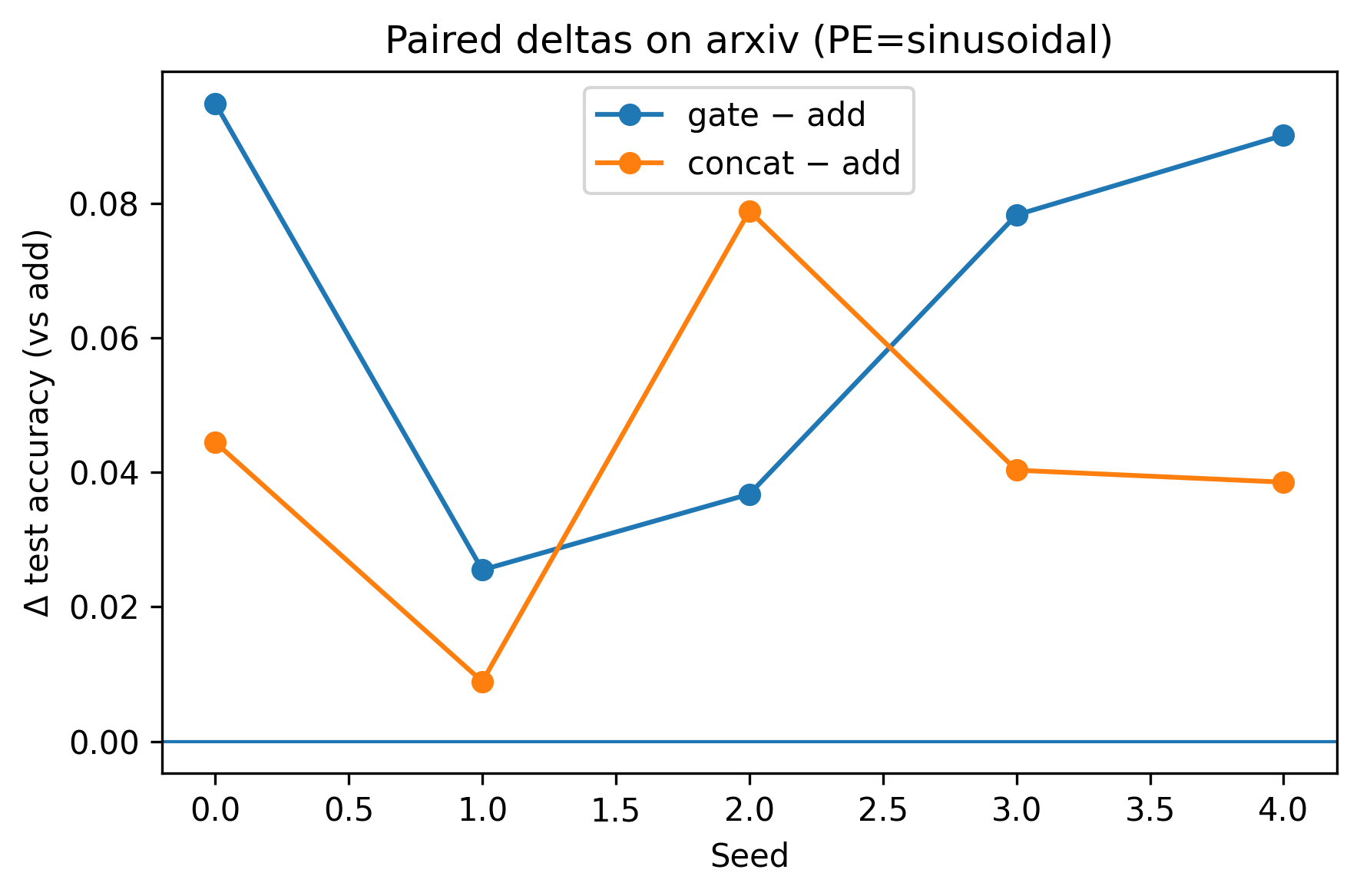}
  \caption{Paired per-seed accuracy deltas (Gate Scalar $-$ Add) on the ArXiv dataset.}
  \label{fig:paired-delta-arxiv}
\end{figure}
To verify that the observed improvements on ArXiv are not due to randomness, we compute paired per-seed accuracy deltas between gated and additive fusion. All seeds exhibit positive deltas, indicating a consistent structural effect.
\subsection{Cross-Dataset Comparison}
We analyze the effect of fusion strategies across datasets that naturally differ in sequence length, rather than post-hoc slicing within a single corpus. AG News contains short texts, IMDB represents medium-length documents, and ArXiv consists of substantially longer inputs.\\
As shown in \textbf{Table \ref{tab:fusion_results}}, fusion choice has negligible and inconsistent impact on AG News and IMDB, where performance saturates and differences across fusion operators fall within statistical variation. Additional paired-seed delta analyses for AG News and IMDB are provided in the \textbf{(Supplementary Experiments ~\ref{sec:Supplementary Experiments})
}. In contrast, on ArXiv—where documents span thousands of tokens—Gate-Scalar fusion consistently outperforms additive fusion, with improvements that are stable across random seeds (\textbf{Figure \ref{fig:paired-delta-arxiv}}).\\
This cross-dataset comparison indicates that the benefit of learnable positional fusion emerges primarily in long-sequence regimes. The absence of gains on shorter datasets suggests that fusion effects are masked when positional structure is simple or local, whereas long documents expose limitations of uniform additive injection.
\section{Local Inductive Bias in Positional Fusion}
This section evaluates whether inductive bias can be introduced at the fusion level without modifying attention.
\begin{table}[H]
    \centering
    \caption{Mean test accuracy and relative inference latency on ArXiv for different fusion mechanisms. Accuracy values are rounded to two decimals; latency is reported qualitatively due to implementation-dependent variance.}
    \label{Mean_acc_inferenceLatency_Arxiv}
    \begin{tabular}{lcc}
        \toprule
        Fusion Mechanism & Mean Test Accuracy & Inference Latency \\
        \midrule
        Add      & $\sim 0.62$ & Lowest \\
        Gate-CNN& $\sim 0.64$& Slightly higher \\
        Gate-Scalar& $\sim \mathbf{0.67}$& Highest \\
        \bottomrule
    \end{tabular}
\end{table}
Gate-Scalar achieves the highest mean accuracy.. Convolutional gating improves over additive fusion, but paired-seed deltas show less consistent sign than Gate, indicating less stable improvements under the current setting. 
\begin{figure}[H]
  \centering
  \begin{subfigure}[t]{0.49\linewidth}
    \centering
    \includegraphics[width=\linewidth]{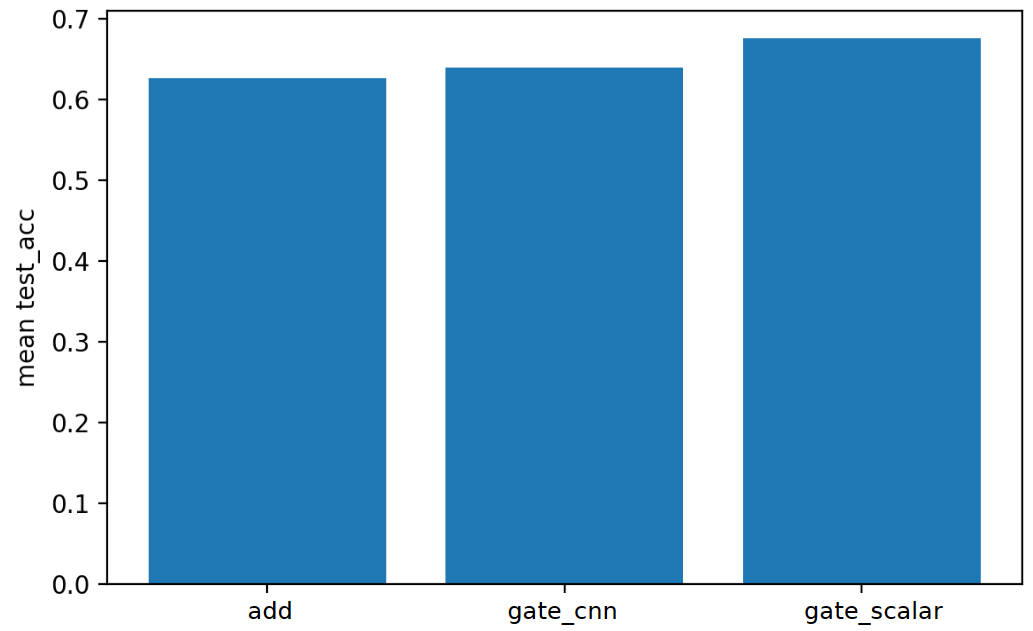}
    \caption{Mean test accuracy on ArXiv by fusion operator.}
    \label{fig:arxiv-acc-fusion}
  \end{subfigure}\hfill
  \begin{subfigure}[t]{0.49\linewidth}
    \centering
    \includegraphics[width=\linewidth]{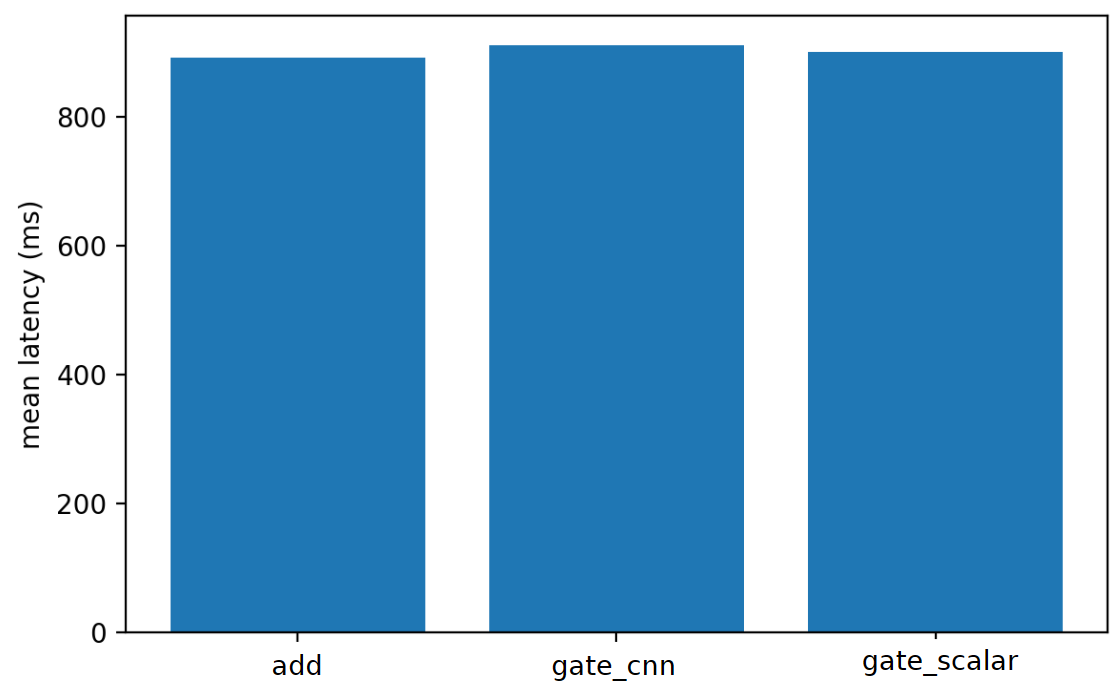}
    \caption{Mean inference latency on ArXiv by fusion operator.}
    \label{fig:arxiv-latency-fusion}
  \end{subfigure}

  \caption{Mean accuracy and inference latency across fusion mechanisms.}
  \label{fig:arxiv-phase2-acc-lat}
\end{figure}
\begin{figure}[H]
  \centering
  \includegraphics[width=0.57\linewidth]{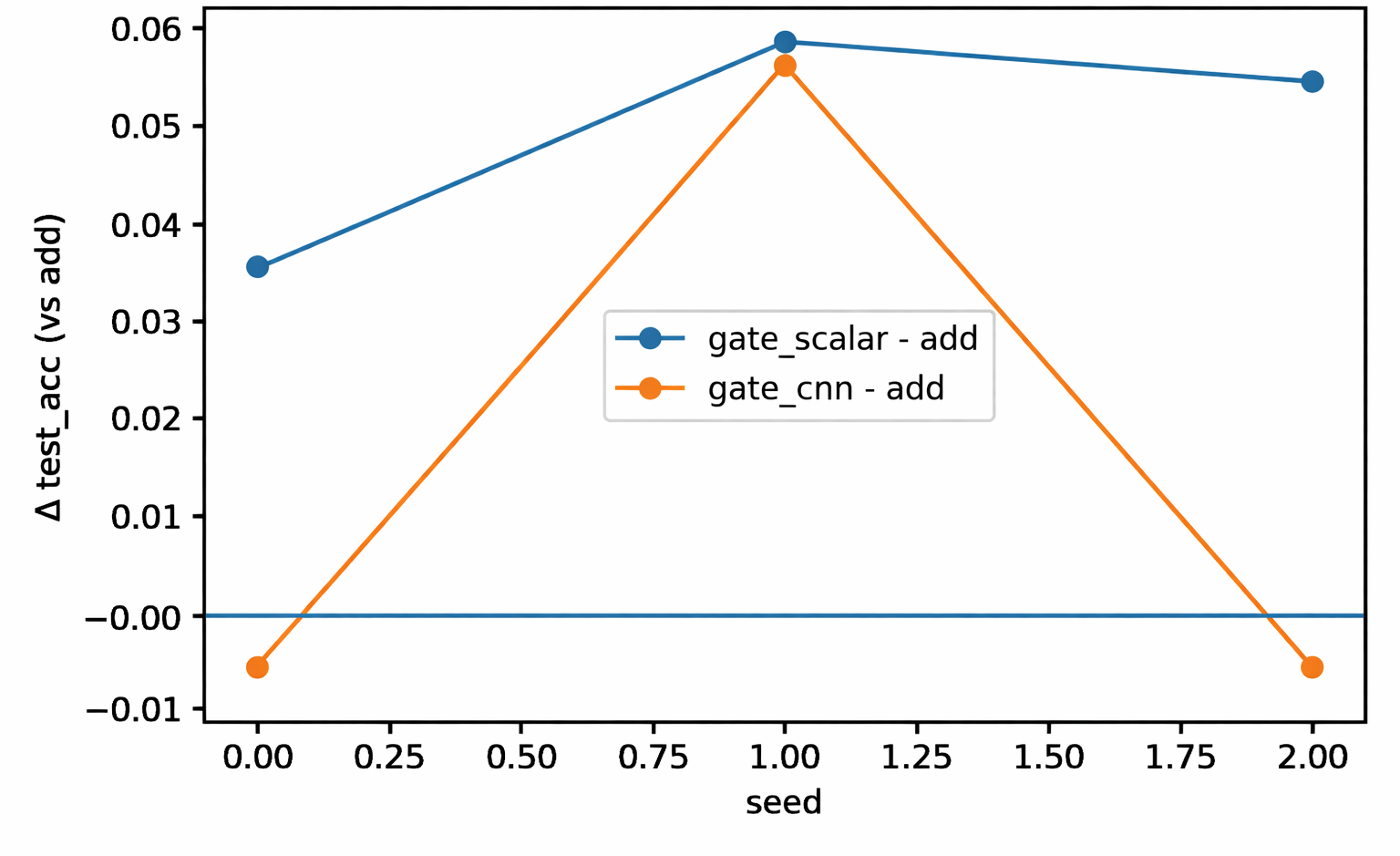}
  \caption{Paired per-seed accuracy deltas on ArXiv relative to Add. Positive values indicate improvement over Add under the same seed.}
  \label{fig:arxiv-paired-deltas}
\end{figure}
Paired-seed deltas show that gated fusion (Gate-Scalar) yields consistently positive improvements relative to additive fusion, whereas convolutional gating exhibits less consistent improvements than Gate-Scalar, indicating less stable gains.
While these results focus on sinusoidal positional encodings, \textbf{Supplementary Experiments \ref{sec:Supplementary Experiments}} reports complementary experiments on the ArXiv dataset showing that the advantage of learnable fusion generalizes across multiple positional encoding families.
\section{Discussion}
Our results show that positional-encoding fusion is not a neutral design choice for long-sequence Transformers. While short-sequence benchmarks saturate and mask fusion effects, long-document classification reveals consistent gains from Gate-Scalar fusion. Cross-dataset comparison across sequence-length regimes shows the effect emerges primarily on long documents. Convolutional gating demonstrates that local inductive bias can be introduced at the fusion level, although its benefits are less stable than those of Gate-Scalar in our experiments. Extended experiments on the ArXiv dataset (reported in the \textbf{(Supplementary Experiments ~\ref{sec:Supplementary Experiments})}) indicate that the advantage of learnable fusion persists across multiple positional encoding families, suggesting that the observed limitation arises from the fusion mechanism rather than from a specific encoding formulation.
\section{Limitations}
This study focuses on classification tasks and a single Transformer architecture. While we evaluated four major positional encoding families on ArXivClassification, future work could extend this analysis to generative tasks, hybrid encodings (e.g., ALiBi), or multimodal settings.
\section{Conclusion}
We have shown that the mechanism used to fuse positional encodings with token embeddings materially affects Transformer performance on long documents. Through controlled experiments, paired-seed analysis and cross-dataset comparison across sequence-length regimes, we demonstrate that gated fusion consistently outperforms the default additive scheme for long-sequence classification. These findings suggest that positional-encoding fusion should be treated as an explicit design choice when building Transformers for long-context tasks.

\section{Supplementary Experiments} \label{sec:Supplementary Experiments}

This supplementary experiment provides additional supporting results beyond the main ArXiv analysis. We first report extended ArXivClassification results across multiple positional encoding families (\textbf{Table 3}) to assess whether fusion effects generalize beyond sinusoidal encodings. We then present paired per-seed accuracy deltas for AG News and IMDB under sinusoidal positional encodings. These short- and medium-length datasets complement the ArXiv findings by showing that fusion effects are small and less consistent across seeds in saturated sequence-length regimes.
\subsection{Cross-Positional Encoding Results on ArXiv }
\begin{table}[h]
    \centering
    \caption{Test accuracy (mean ± standard deviation) on the long-sequence ArXiv classification dataset across four positional encoding families and four fusion strategies. Results are averaged over five random seeds using identical architectures and hyperparameters, where MLP denotes a content–position fusion baseline evaluated \textbf{only on ArXiv}.}
    \label{tab:placeholder_label}
    \begin{tabular}{lcccc}
        \toprule
        Positional Encoding & Add & Concat & Gate-Scalar& MLP-Gate\\
        \midrule
        \textbf{Sinusoidal}       & 59.22 $\pm$ 1.07 & 63.44 $\pm$ 2.27 & 65.73 $\pm$ 4.01 & 65.50 $\pm$ 2.31 \\
        \textbf{Learned Absolute} & 62.29 $\pm$ 3.24 & 60.83 $\pm$ 2.77 & 64.61 $\pm$ 2.83 & 64.28 $\pm$ 5.08 \\
        \textbf{RoPE}             & 58.47 $\pm$ 2.58 & 64.55 $\pm$ 1.82 & 65.61 $\pm$ 2.81 & 64.33 $\pm$ 3.67 \\
        \textbf{Relative}         & 62.48 $\pm$ 2.53 & 59.83 $\pm$ 2.11 & 65.55 $\pm$ 3.15 & 65.23 $\pm$ 2.34 \\
        \bottomrule
    \end{tabular}
\end{table}
Across all four positional encoding families, Here, Gate-Scalar refers to the scalar positional gate used in the main paper, while Gate-MLP denotes a feature-wise MLP-based gating variant evaluated only in Supplementary Experiments \ref{sec:Supplementary Experiments}. Neither gating variant dominates universally: scalar gating performs slightly better for sinusoidal and RoPE encodings, while MLP-based gating achieves comparable or marginally higher accuracy for learned absolute and relative encodings. These results indicate that the primary source of improvement lies in introducing a learnable fusion mechanism, rather than in a specific gating parameterization or positional encoding formulation. 
\subsection{Short- and Medium-Length Dataset Analysis}
\subsubsection{AG News}
\begin{figure}[H]
  \centering
  \includegraphics[width=0.57\linewidth]{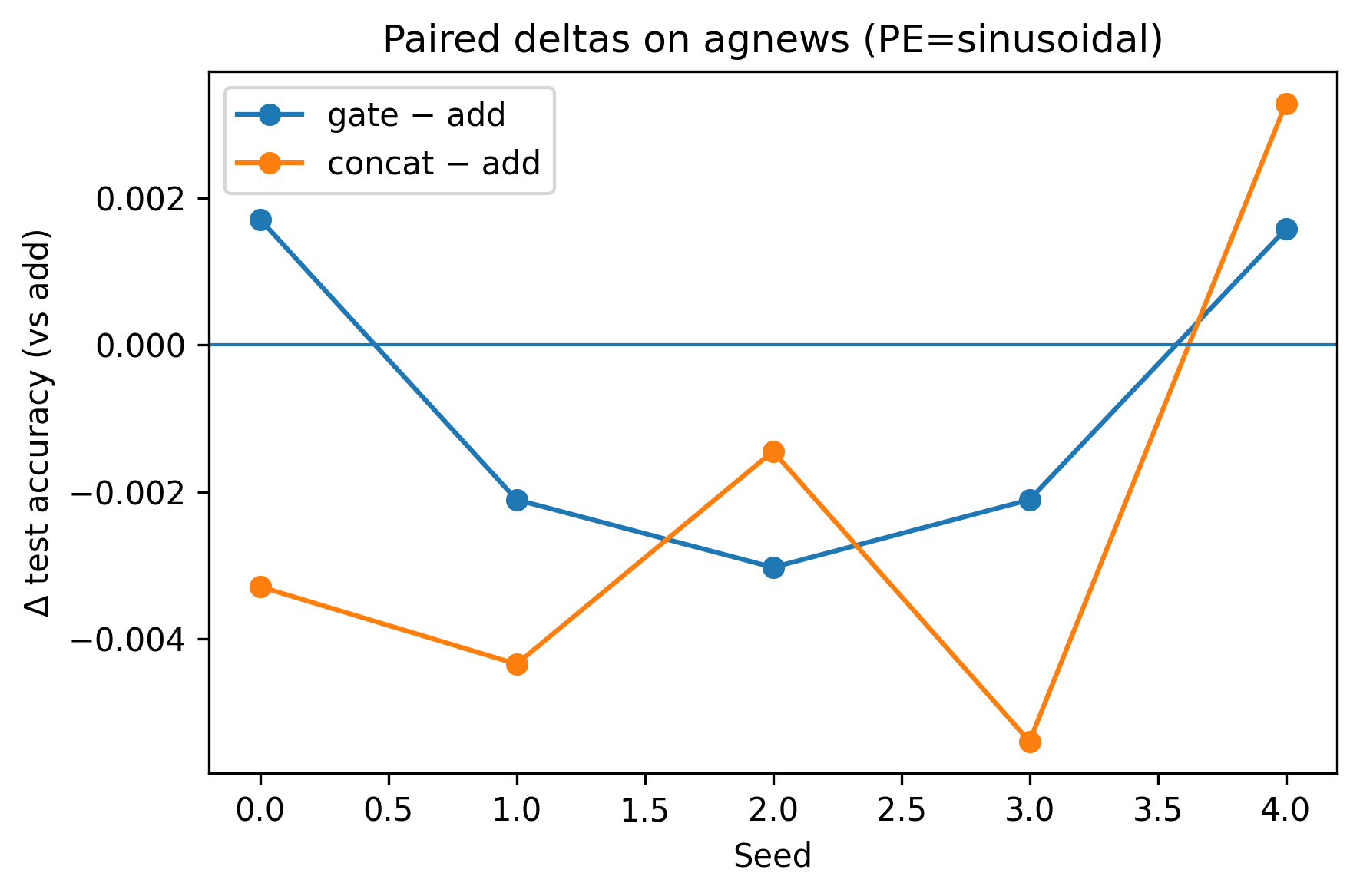}
  \caption{Paired per-seed accuracy deltas (Gate Scalar - Add) on AG News using sinusoidal positional encoding. Deltas are small may change sign across seeds, consistent with saturation effects on short sequences}
  \label{fig:AG_News-paired-deltas}
\end{figure}
\textbf{Figure \ref{fig:AG_News-paired-deltas} }shows paired per-seed accuracy deltas between gated fusion and additive fusion on the AG News dataset. The deltas are small in magnitude and exhibit sign changes across seeds. This indicates that, for short sequences, the choice of fusion operator does not produce a stable or systematic advantage, and performance differences fall within statistical variation.

\subsubsection{IMDB}
\begin{figure}[H]
  \centering
  \includegraphics[width=0.57\linewidth]{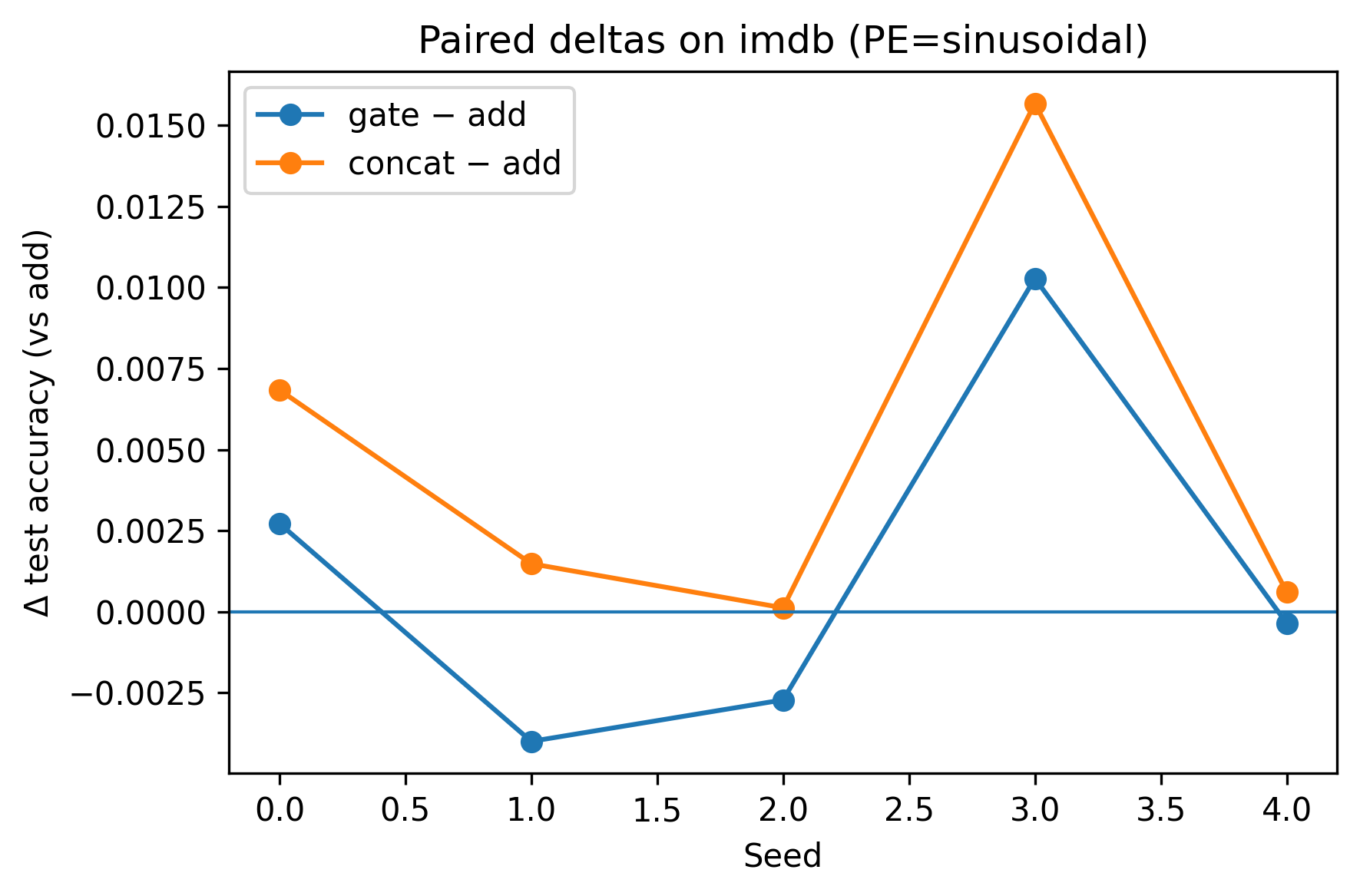}
  \caption{Paired-per-seed accuracy deltas (Gate Scalar - Add) on IMDB using sinusoidal positional encoding. Deltas are small and less sign-consistent than on ArXiv, consistent with limited differentiation between fusion strategies on medium-length sequences.}
  \label{fig:IMDB-paired-deltas}
\end{figure}
\textbf{Figure \ref{fig:IMDB-paired-deltas} }presents paired per-seed accuracy deltas for the IMDB dataset. Similar to AG News, the deltas are modest and less sign-consistent than those observed on ArXiv. This suggests that, for medium-length documents, fusion mechanisms offer limited separation and do not yield robust improvements across random seeds.

\subsection*{Summary of Supplementary Experiments Findings}
Taken together, the supplementary experiments' results reinforce the main conclusions of the paper. Learnable fusion mechanisms provide clear benefits on long-document classification, and this effect generalizes across multiple positional encoding families. In contrast, on short and medium-length datasets, fusion effects are small and unstable, consistent with performance saturation in these regimes.

\bibliographystyle{unsrtnat}  

 \bibliography{references}

\end{document}